\documentclass[runningheads]{llncs}
\usepackage{pdfpages}
\usepackage[width=122mm,left=12mm,paperwidth=146mm,height=193mm,top=12mm,paperheight=217mm]{geometry}
\usepackage{graphicx,amsmath,amssymb,subcaption,cite,floatrow}
\usepackage[colorlinks=true,bookmarks=false]{hyperref}

\newcommand{\app}{\raise.17ex\hbox{$\scriptstyle\sim$}}

\newlength\savewidth

\begin{document}
\pagestyle{headings}\mainmatter
\title{Parallel Convolutional Networks for Image Recognition via a Discriminator}

\institute{School of Automation, Huazhong University of Science and Technology
\email{albert\_yang@hust.edu.cn, penggang@hust.edu.cn}
}
\titlerunning{D-PCN}
\author{Shiqi Yang, Gang Peng}
\authorrunning{Shiqi Yang, Gang Peng}
\maketitle

\begin{abstract}
	In this paper, we introduce a simple but quite effective recognition framework dubbed D-PCN, aiming at enhancing feature extracting ability of CNN. The framework consists of two parallel CNNs, a discriminator and an extra classifier which takes integrated features from parallel networks and gives final prediction. The discriminator is core which drives parallel networks to focus on different regions and learn different representations. The corresponding training strategy is introduced to ensures utilization of discriminator. We validate D-PCN with several CNN models on benchmark datasets: CIFAR-100, and ImageNet, D-PCN enhances all models. In particular it yields state of the art performance on CIFAR-100 compared with related works. We also conduct visualization experiment on fine-grained Stanford Dogs dataset to verify our motivation. Additionally, we apply D-PCN for segmentation on PASCAL VOC 2012 and also find promotion.
	
\end{abstract}

\section{Introduction}
\label{sec1}

Since the AlexNet~\cite{krizhevsky2012imagenet} sparked off the passion for research on convolutional neural networks (CNNs), CNNs have been improving the performance of image classification continuously. And heterogeneous successive brilliant CNN models lead this wave with compelling results, besides, state of the art of various vision tasks, such as detection~\cite{ren2015faster,redmon2016you,liu2016ssd} and segmentation~\cite{long2015fully,chen2017deeplab}, is advancing rapidly leveraging the power of CNN.

\begin{figure}[!tb]\centering
	\includegraphics[width=1\textwidth]{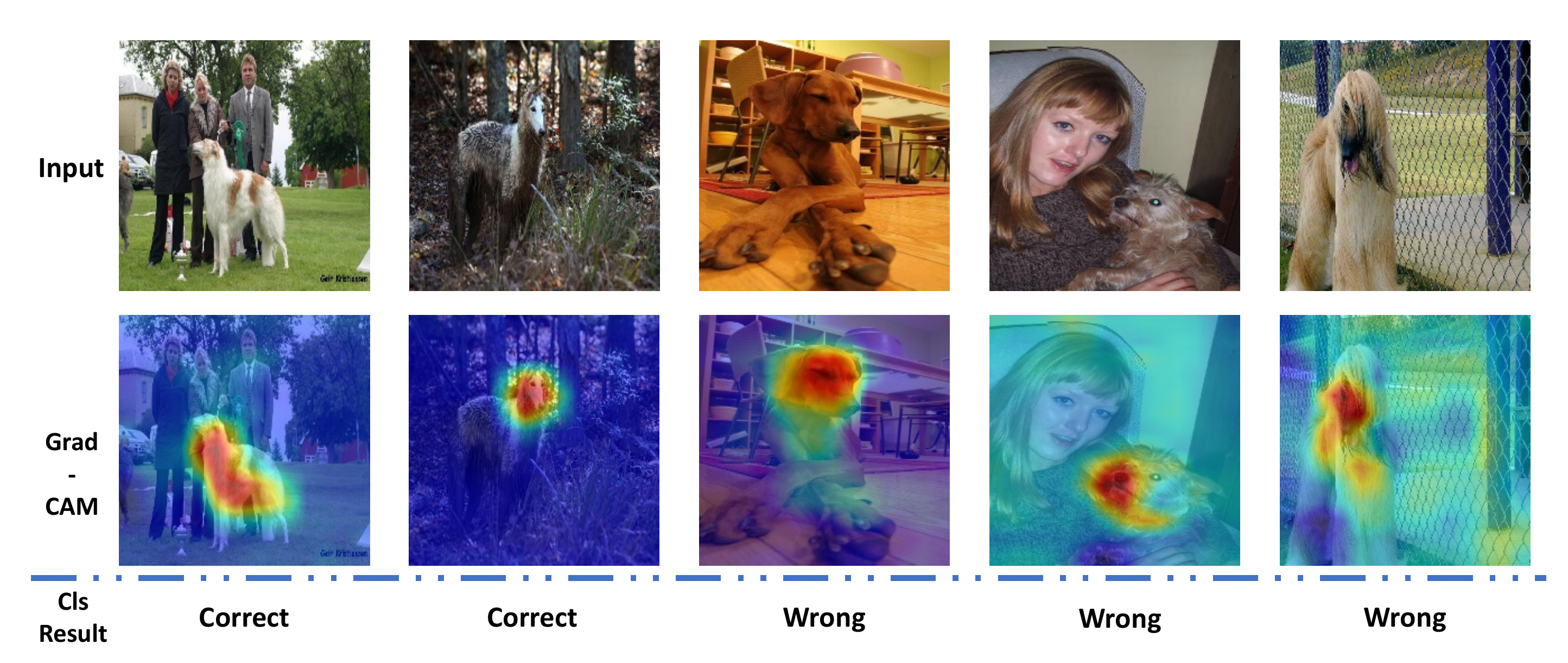}
	\caption{Grad-CAM visualization of VGG16, which aims to distinguish diverse categories of dogs. The second row shows the class-discriminative regions localized by CNN. The cls result means whether network predicts correctly.}
	\label{fig1}
\end{figure}

A number of recent papers~\cite{zeiler2014visualizing,Zhou2015,zhou2016learning,bau2017network,selvaraju2016grad} have tried to lend insights on interpretability of CNN. These methods focus on understanding CNN by visualizing learned representations. An interesting conclusion~\cite{Zhou2015,zhou2016learning} has been drawn that CNN has ability to localize objects without any supervision in classification task. As shown in Figure~\ref{fig1}, we visualize the VGG16 using Grad-CAM~\cite{selvaraju2016grad}. We posit that single network may not notice all informative regions or details which leads to misclassification as exhibited in Figure~\ref{fig1}, meaning focusing on specific areas since some different categories may have these regions in similar. Based on this point, in this paper we propose a parallel networks architecture dubbed D-PCN, which coordinates parallel networks to achieve diverse representations under the guide of a discriminator. The final prediction is reported by the extra classifier. We adopt a training method which is modified from adversarial learning to achieve our goal.

We implement D-PCN on CIFAR-100~\cite{krizhevsky2009learning}, ImageNet\cite{chrabaszcz2017downsampled,deng2009imagenet} datasets with NIN~\cite{lin2013network}, ResNet~\cite{he2016deep}, WRN~\cite{zagoruyko2016wide}, ResNeXt~\cite{xie2016aggregated} and DenseNet~\cite{huang2017densely}. In experiments, the performance of D-PCN ascends greatly compared with single base CNN, and it's proved that performance improvement is not merely from more parameters. In particular, our method has outperformed all advanced related approaches which use multiple subnetworks on CIFAR-100. We also apply Grad-CAM~\cite{selvaraju2016grad} to visualize D-PCN with VGG16~\cite{simonyan2014very} on a fine-grained classification dataset, Stanford Dogs~\cite{khosla2011novel}, and the result verifies our motivation. In addition, we introduce D-PCN into FCN~\cite{long2015fully} on PASCAL VOC 2012 segmentation task, and experiment result demonstrates that D-PCN enhances the network. 

We summarize our contributions as follows:
\begin{itemize}
	\item We propose the D-PCN, a simple but quite effective framework to enhance feature extracting ability of CNN, which outperforms other related methods.
	\item Two parallel networks in D-PCN focus on different regions of input respectively, that leads to more discriminative representations after features fusion.
	\item We propose a novel training method, and it's of high efficiency to be applied for D-PCN.
\end{itemize}

\section{Related Work}
\label{sec2}

CNN based models occupy advanced performance in almost all computer vision areas, including classification, detection and segmentation. Many attempts have been made to design efficient CNN architecture. In early stage, the networks from AlexNet~\cite{krizhevsky2012imagenet} to VGGnet~\cite{simonyan2014very} tend to get deeper, and thanks to skip connection, ResNet~\cite{he2016deep} can contain extreme deeper layers. WRN~\cite{zagoruyko2016wide} demonstrates the fact that increasing width can improve performance too. And there also exist other innovative designed models, such as Inception~\cite{szegedy2015going}, ResNeXt~\cite{xie2016aggregated}, DenseNet~\cite{huang2017densely}, which improve capability of CNN further. Besides, many new modules have been constructed. For example, serials of activation functions have been introduced, like PReLU~\cite{he2015delving} which accelerates convergence, and the interesting SeLU~\cite{klambauer2017self}. Additionally, batch normalization~\cite{ioffe2015batch} is used to normalize input of layers and improve performance. Some other works achieve enhancement by employing regularizer such as dropout~\cite{srivastava2014dropout} and maxout~\cite{goodfellow2013maxout}. 

Although all these methods turn out to be very helpful, but sometimes designing new network models or activation units is of high complexity. Speculating on how to strengthen ability of existed CNN models is a feasible approach. Several works have paid attention to that, including Bilinear CNN~\cite{lin2015bilinear}, HD-CNN~\cite{yan2015hd}, DDN~\cite{murthy2016deep} and DualNet~\cite{Hou2017DualNet}, all of which resort to multiple networks. HD-CNN~\cite{yan2015hd} embeds CNN into two-level category hierarchy, it separates easy classes with a coarse category classifier while distinguishing different classes using a fine classifier. And DDN~\cite{murthy2016deep} automatically builds a network that splits the data into disjoint clusters of classes which would be handled by the subsequent expert networks.

Bilinear CNN~\cite{lin2015bilinear} and DualNet~\cite{Hou2017DualNet} are more related to our work which all use parallel networks. But our work is distinctive from them. In Bilinear CNN~\cite{lin2015bilinear} the parallel networks have different parameters numbers and receptive fields, and in fact Bilinear CNN is eventually implemented with a single CNN using weights sharing, while D-PCN has two identical networks. DualNet~\cite{Hou2017DualNet} is the first to focus on the cooperation of multiple CNNs. Although it shares same philosophy with us which means deploying identical parallel networks, it puts an extra classifier in the end to participate in joint training with parallel networks. The extra classifier guarantees divergence of two networks in DualNet, and final prediction is a weighted average over three classifiers. While D-PCN uses a discriminator to drive two networks to learn different representations, and the added extra classifier doesn't take part in training with parallel. Moreover, the final prediction in D-PCN is reported by extra classifier alone. Besides, the motivation is different, two subnetworks in D-PCN are expected to localize distinctive regions of input. A novel training strategy adapted from adversarial learning is proposed to achieve it in D-PCN. Additionally, D-PCN is much easy to implement compared with related works.

\section{D-PCN}
\subsection{Motivation}
\label{sec3.1}
\begin{figure}[!tb]\centering
	\includegraphics[width=1\textwidth]{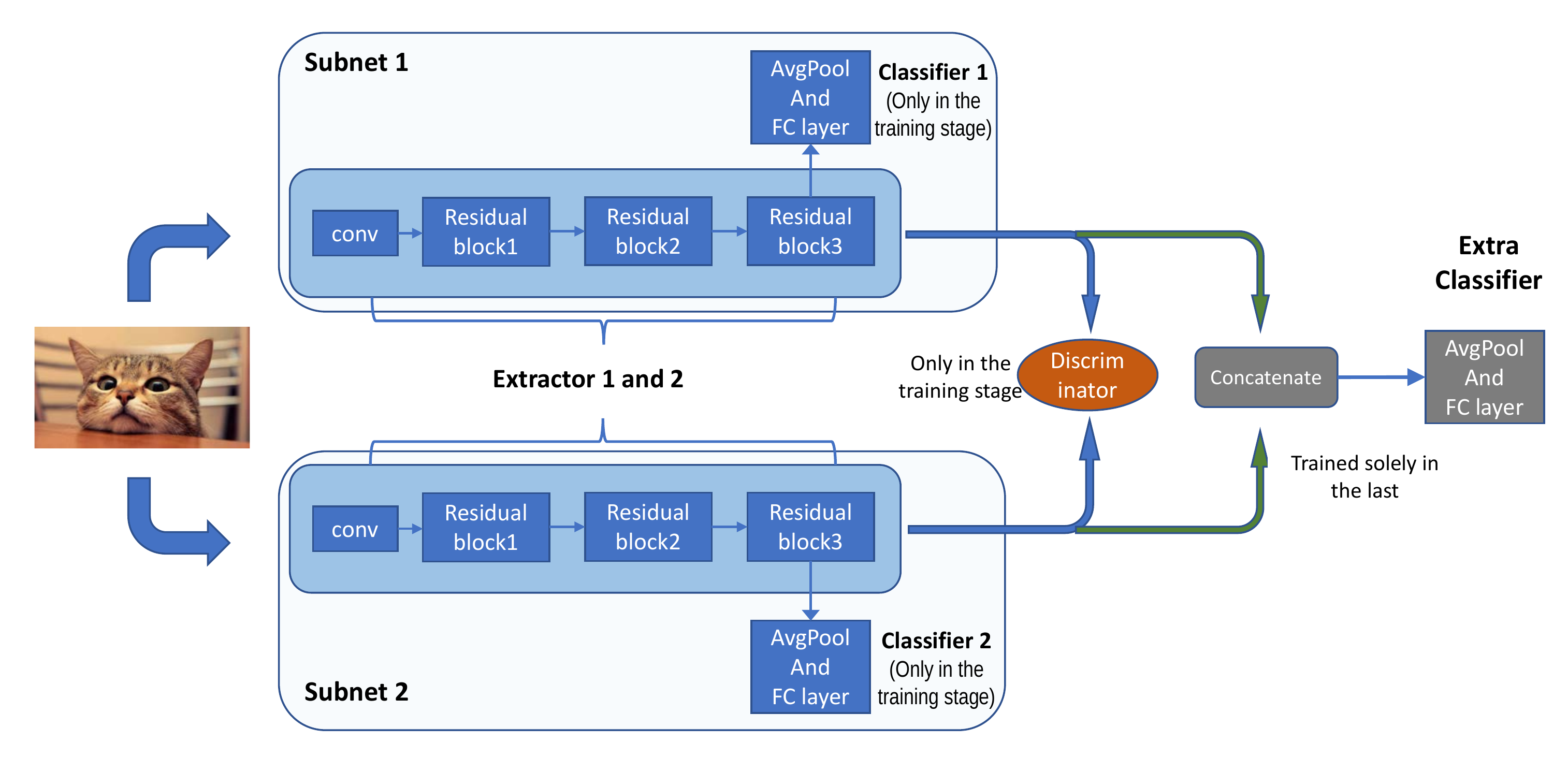}
	\caption{\textbf{The architecture of D-PCN based on ResNet-20. }Noted that classifier of two networks and the discriminator just appear in training process.}
	\label{fig3}
\end{figure}
Nowadays, neural networks are still trained with back propagation to optimize the loss function, the process is driven by losses generated at higher layers. Consequently, as demonstrated in \cite{Hou2017DualNet}: In the optimization of single network, some distinctive details of the objects, which are low-level but essential to discriminate the classes of strong similarity, are likely to be dropped in the middle layers or overwhelmed by massive useless information, since the loss signals received by shallow layers for parameter update have been filtered by multiple upper layers. And these may happen constantly in whole propagation process when loss signals flow from higher to lower layers. All in all, it is tough for a single network to extract all details of input.

As mentioned in Section~\ref{sec1}, there are various intriguing works for visualization on CNN lately, which point out that CNN can localize related target object in a spontaneous way. We conjecture that the missing of some information in the optimizing process~\cite{Hou2017DualNet} may lead to inaccurate localization and misclassification as shown in Figure~\ref{fig1}. We want to utilize this characteristic of CNN to improve performance of vision tasks. Therefore we hope to find a way to compel multiple networks to focus on different regions or details, which implies one network can learn features omitted by others.

Recently, Generative Adversarial Nets (GAN)~\cite{goodfellow2014generative} are prevalent. In a GAN, the discriminator and generator are playing a max-min game which is realized by adversarial learning. This competition between them can drive both teams to improve their methods until the spurious are indistinguishable from the genuine ones. The adversarial learning is depicted as below:
\[
\begin{split}
\underset{G}{min}\ \underset{D}{max}\ V(D,G)=\mathbb{E}_{x\sim p_{data}(\textbf{x})}[logD(\textbf{x})]
+\mathbb{E}_{z\sim p_{\textbf{z}}(\textbf{z})}[log(1-D(G(\textbf{z})))]
\end{split}
\]
$D$ here represents discriminator and $E$ means generator.

Arguably, the discriminator can be regarded as a relay through which generator acquires information from real input, meanwhile it differentiates generated one and real one. ~\cite{goodfellow2014generative} reformulates $\underset{D}{max}\ V(D,G)$ as:
\[
\begin{split}
\underset{D}{max}\ V(D,G)= -log4 + 2\cdot JSD(p_{data}||p_{g})
\end{split}
\]
$JSD$ means Jensen-Shannon divergence. The generator is to minimize the divergence so as to generate indistinguishable object.

Inspired by it, we propose a parallel networks framework named D-PCN by transforming adversarial learning to a maximization optimization problem. In D-PCN, there are two identical parallel networks, a discriminator, and an extra classifier giving final prediction. The key component of D-PCN is the discriminator which can coordinate parallel networks to learn features from different regions or aspects. It's achieved by a training strategy as below:
\begin{multline}
	\underset{E_1,E_2}{max}\ \underset{D}{max}\ V(D,E_1,E_2)=\mathbb{E}_{x\sim input}[logD(E_1(x))]
	+\mathbb{E}_{x\sim input}[log(1-D(E_2(x)))]
	\label{eq1}
\end{multline}
where the $E_1,E_2$ symbolize extractors of subnetwork 1 and 2 respectively, equal to generator in GAN. And $E_1(x),E_2(x)$ means features learned by networks. The equation~\ref{eq1} is to enforce two subnetworks to learn two different features spaces. Unlike GAN, we want to maximize $\underset{D}{max}\ V(D,E_1,E_2)= -log4 + 2\cdot JSD(p_{E_1(x)}||p_{E_2(x)})$ to expand distribution distance between two extractors, by which means subnetworks can learn different features. Though there maybe a solution to equation~\ref{eq1} where the ordering of features are perturbed between $E_1(x)$ and $E_2(x)$, we posit that since we force subnets to achieve good performance on classification meanwhile, the perturbation (even very minor) between $E_1(x)$ and $E_2(x)$ will always exist, which will result in extra information learned. This will be proved in Section~\ref{vi}.

\subsection{Architecture}
\label{sec3.2}
In D-PCN, there are two subnetworks with same architecture. Two subnetworks can be replaced with any present CNN model.

In order to articulate the framework, we take a D-PCN based on ResNet-20~\cite{he2016deep} for example, which is presented in Figure~\ref{fig3}. We separate a single network into an extractor and a classifier, corresponding to the figure. The classifier merely contains a pool layer and a fully connected (\emph{fc}) layer, the other lower layers belong to extractor. The discriminator is comprised by several convolutional layers, with batch normalization and Leaky ReLU~\cite{maas2013rectifier}. Noted that in experiments sigmoid activation is added in the end of discriminator specially for D-PCN based on NIN~\cite{lin2013network}.

The reasons why we choose features of higher layer to be sent to discriminator lie in two aspects. First, CNN extracts hierarchical representations from edges to almost entire object with encoded features~\cite{zeiler2014visualizing}, so the features in high layers are much discriminative. Discriminator which takes in these features can acquire enough information to guide training. Second, the feature size in higher layers is much small, and this will reduce computational cost. All other D-PCNs with various CNN models adopt similar position as division of extractor and classifier. Just as shown in Figure~\ref{fig3}, during training procedure, extractors along with their own classifiers and discriminator will get trained in the beginning. Then features will be integrated and input to extra classifier, which will be trained solely. In inference stage, there are no discriminator and classifier in both subnetworks, and final prediction is reported by extra classifier.

For simplicity, we adopt concatenating as fusing method. After features integrated, the whole framework can obtain more discriminative representations. The extra classifier keeps same architecture as classifiers in subnetworks, but with width doubling because of concatenating. By the way, since NIN just has a pool layer in the end for prediction, we add a \emph{fc} layer in extra classifier to maintain proper structure.

\subsection{Training Method}
\label{sec3.3}

\begin{figure}[!tb]\centering
	\includegraphics[width=0.85\textwidth]{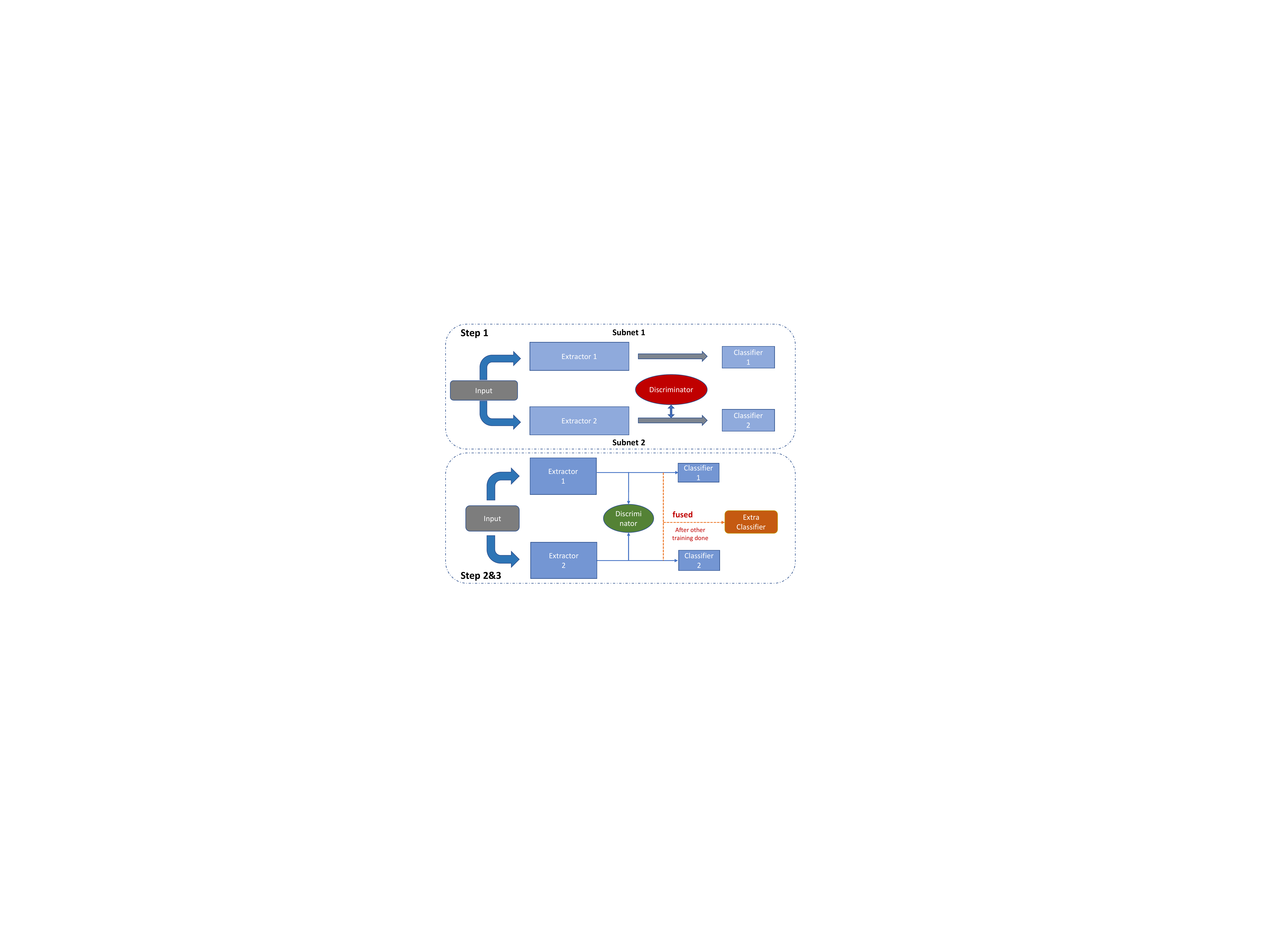}
	\caption{\textbf{The training strategy of D-PCN. }At first, parallel networks start being trained, the loss of discriminator is attached to one of them to realize different parameters of two networks. Then two subnetworks are trained jointed in accompany with discriminator. In completion of joint training, extra classifier will be trained with fused features from two extractors.}
	\label{fig4}
\end{figure}

The proposed training method is crucial to coordinate parallel networks to localize diversely and learn different features. The process contains three steps. The discriminator in D-PCN works like a binary classifier, and it tells which network the features are from, and the loss signal it spreads to parallel networks will encourage one network to learn features omitted by another. Since there is no constraint to make feature of subnets complementary totally, duplications will exist in representations of subnets, but it's important that subnets catch different details, which will be proved in Section.~\ref{vi}. Unlike using iterative training in DualNet~\cite{Hou2017DualNet}, we deploy a 3-step training, illustrated in Fig.~\ref{fig4}.

\subsubsection{Training Step 1}
Because it's tough for parallel networks with same values of parameters to converge by our joint training method, we need to initialize subnetworks with different weights. Since we have a discriminator in D-PCN, we opt to make full use of it. In this stage, discriminator is initialized and fixed, and the loss value from discriminator is added to one of the subnetwork, by which means features learned by parallel networks can be discriminative and distinctive simultaneously. For subnetwork 1, the loss function is defined as:
\begin{eqnarray}
&&L_1=L_{cls_1}
\end{eqnarray}
while loss function of another is defined as:
\begin{eqnarray}
L_2=L_{cls_2}+\lambda L_{D_2}\\
L_{D_2}=\frac{1}{n}\sum^n{{[D(E_2(input))]^2}} 
\end{eqnarray}
The $L_{cls}$ means cross entropy loss for classification, and $L_{D_2}$ is a L2 loss from discriminator. After a few epoches, Step 1 is finished.

\subsubsection{Training Step 2}
Joint training based on Equation~\ref{eq1} starts. Loss function of subnetwork 1 is defined as:
\begin{eqnarray}
L_1=L_{cls_1}+\lambda L_{D_1}\\
L_{D_1}=\frac{1}{n}\sum^n{{[1-D(E_1(input))]^2}}
\end{eqnarray}
In the meantime the corresponding one of subnetwork 2 is defined as:
\begin{eqnarray}
L_2=L_{cls_2}+\lambda L_{D_2}\\
L_{D_2}=\frac{1}{n}\sum^n{{[D(E_2(input))]^2}}
\label{eq8}
\end{eqnarray}
As for discriminator, it follows the paradigm of GAN:
\begin{equation}
L_D=L_{D_1}+L_{D_2}
\end{equation}

Extractors in parallel networks can be regarded as counterparts of generator in GAN. In above training, $L_{cls}$ ensures that learned features are discriminative, meanwhile $L_{D_1}$, $L_{D_2}$ make sure that features from subnetworks are different. By the way, $L_{D_1}$ and $L_{D_2}$ both can be seen as regularization to some extend.

\subsubsection{Training Step 3}
In this stage, we remove classifiers in two-stream networks, so does discriminator. Features from two networks get integrated and are sent to extra classifier. All extractors are fixed, and we only train extra classifier.

In training, $\lambda$ is set to 1, and we find it's sufficient to promote performance of CNN significantly. We emphasize that discriminator receives discriminative features from subnetwork 1 and can instruct the training of subnetwork 2 and vice verse, although the process is operated in class level. In addition, our method is much easy to implement compared with related works.

\section{Experiments}
In this section, we empirically demonstrate the effectiveness of D-PCN with several CNN models on various benchmark datasets and compare it with related state of the art methods. Additional experiments for visualization and segmentation are also conducted. All experiments are implemented with PyTorch\footnote{http://pytorch.org/} on a TITAN Xp GPU.
\subsection{Classification results on CIFAR-100}
\begin{table}[tb]
	\caption{\textbf{Compared with recent related works on CIFAR-100 without data augmentation.} The accuracy means the top-1 accuracy on CIFAR-100 test datasets. *-with data augmentation and 10 view testing.
	}
	\begin{center}
		\footnotesize
		\setlength{\tabcolsep}{5pt}
		\begin{tabular}{|l|c|}
			\hline
			Method& Test Accuracy\\
			\hline
			Maxout Network~\cite{goodfellow2013maxout}  & 61.43\% \\
			Tree based priors~\cite{srivastava2013discriminative} & 63.15\% \\
			Network in Network~\cite{lin2013network} & 64.32\% \\
			DSN~\cite{lee2015deeply} & 65.43\% \\
			NIN+LA units~\cite{agostinelli2014learning} & 65.60\% \\
			HD-CNN*~\cite{yan2015hd} & 67.38\% \\
			DDN~\cite{murthy2016deep} & 68.35\% \\
			DNI, DualNet~\cite{Hou2017DualNet} & 69.76\% \\
			\hline
			\hline
			\textbf{D-PCN (ours)}& \textbf{71.10\%} \\
			\hline
		\end{tabular}
	\end{center}
	\label{tab1}
\end{table}

The CIFAR-100~\cite{krizhevsky2009learning} dataset consists of 100 classes and total 60000 images with 32x32 pixels each, in which there are 50000 for training and 10000 for testing. We simply apply normalization for images using means and standard deviations in three channels.

For convenience of making comparison with related works, which use NIN~\cite{lin2013network} as base model, such as HD-CNN~\cite{yan2015hd}, DDN~\cite{murthy2016deep}, DualNet~\cite{Hou2017DualNet}, we build a D-PCN based on NIN. We follow the setting of NIN in~\cite{murthy2016deep,Hou2017DualNet}, and D-PCN is trained without data augmentation. Table~\ref{tab1} shows performance comparisons between several works. Noted directly compared to D-PCN are~\cite{lin2013network,yan2015hd,murthy2016deep,Hou2017DualNet}, which are all built on NIN and deploying multiple networks. And HD-CNN actually uses cropping and 10 view testing~\cite{krizhevsky2012imagenet} as data augmentation, but it's a representative work using multiple subnetworks, for which reason it's listed here. To the best of our knowledge, DualNet reports highest accuracy on CIFAR-100 without augmentation before D-PCN. Our work surpasses DualNet by 1.34\%.

Furthermore, we make several elaborate comparisons between D-PCN and DualNet in order to prove the effectiveness of our work. As shown in Table~\ref{tab2}, we list all prediction results in DualNet and D-PCN. DualNet consists of two parallel networks and an extra classifier, final prediction is given by a weighted average of all three classifiers, while ours is provided by extra classifier alone. For ResNet, DualNet takes additional data augmentation approaches, which may explain why accuracy of base network in DualNet overtakes ours. However, D-PCN still outperforms DualNet except for ResNet-20. The accuracy of two subnetworks in D-PCN already exceeds base network. It's worth nothing that the extra classifiers achieve significant boost over base single network after features integration. These promising results may signify representations learned by parallel networks in D-PCN are indeed different.

\begin{table}[tb]
	\caption{\textbf{Comparison between DualNet and our D-PCN on CIFAR-100. } Here we report predictions from all classifiers in two frameworks. The top half shows the results of DualNet, in which the \emph{classifier average} means the weighted average of all three classifiers in DualNet. The bottle half shows the results of our D-PCN. * means that DualNet uses changing contrast, brightness and color shift as additional data augmentation ways, while + means our D-PCN only adopts cropping randomly with padding.
	}
	\begin{center}
		\footnotesize
		\setlength{\tabcolsep}{0pt}
		\begin{tabular}{|l|c|c|c|c|} 
			\hline
			Method &&&&  \\
			\hline
			\hline
			\textbf{DualNet}~\cite{Hou2017DualNet}  & \textbf{NIN} & \textbf{ResNet-20*} & \textbf{ResNet-34*}& \textbf{ResNet-56*} \\
			\hline
			\textbf{base network}  & 66.91\% &\emph{69.09\%}  & \emph{69.72\%} & \emph{72.81\%} \\
			\hline
			\emph{iter training (Extra Classifier)}  & 69.01\%   & 71.93\% & 73.06\% & 75.24\% \\
			\hline
			\emph{iter training (classifier average)}& 69.51\% &	 72.29\%   & 73.31\% & 75.53\% \\
			\hline
			\emph{joint finetuning (classifier average)}& \textbf{69.76\%}  & \textbf{72.43\%}  & \textbf{73.51\%} & \textbf{75.57\%}\\
			\hline
			\hline
			\hline
			\textbf{D-PCN} & \textbf{NIN} & \textbf{ResNet-20+} & \textbf{ResNet-34+}  & \textbf{ResNet-56+}\\
			\hline
			\textbf{base network}  & 66.63\%  & 67.89\%   & 68.70\%  & 71.98\% \\
			\hline
			\emph{classifier of subnet1} & 68.03\% & 68.15\%   & 69.76\%   & 73.44\%\\
			\hline
			\emph{classifier of subnet2} & 67.96\% & 68.69\%   & 69.94\%   & 74.01\%\\
			\hline
			\emph{extra classifier} & \textbf{71.10\%} & \textbf{72.39\%}  & \textbf{74.07\%} & \textbf{76.39\%}  \\
			\hline
		\end{tabular}
	\end{center}
	\label{tab2}
\end{table}

Moreover, we also evaluate other celebrated CNN models, including WRN~\cite{zagoruyko2016wide}, DenseNet~\cite{huang2017densely} and ResNeXt~\cite{xie2016aggregated}. The results are shown in Table~\ref{tab3}. We can find D-PCN improve the accuracy of all these models.

\begin{table}[!tb]
	\caption{\textbf{Accuracy of D-PCNs based on several models on CIFAR-100. }All results are run by ourselves and produced with cropping randomly implemented as the only data augmentation method.
	}
	\begin{center}
		\footnotesize
		\setlength{\tabcolsep}{9pt}
		\begin{tabular}{|l|c|c|c|} 
			\hline
			\textbf{D-PCN} & \textbf{DenseNet-40} & \textbf{WRN-16-4} & \textbf{ResNeXt-29,8x64d} \\
			\hline
			\textbf{base network}  & 70.03\%  & 76.72\%  & 81.77\%  \\
			\hline
			\emph{classifier of subnet1} & 70.33\% & 77.63\%   & 81.98\%  \\
			\hline
			\emph{classifier of subnet2} & 70.10\% & 77.76\%   & 82.41\%  \\
			\hline
			\emph{extra classifier} & \textbf{71.43\%} & \textbf{80.19\%}  & \textbf{84.59\%}  \\
			\hline
		\end{tabular}
	\end{center}
	\label{tab3}
\end{table}

\textbf{Analysis of D-PCN on ResNet and DenseNet }Just as discussed in DPN~\cite{chen2017dual}, ResNet tends to reuse the feature and fails to explore new ones while DenseNet is able to extract new features. In short, DenseNet can learn comprehensive features as much as possible. And results in Table~\ref{tab2} and Table~\ref{tab3} reflect these characteristics, where D-PCN can bring more promotions for ResNet than DenseNet.

\textbf{Compared with model ensemble }For sake of further verifying effectiveness of D-PCN, we also report results of model ensemble. Specifically, we train two CNN models independently, initialized with normal distribution or using Xavier~\cite{glorot2010understanding} and Kaiming~\cite{he2015delving} initialization, and final prediction is obtained with their predictions averaged. As illustrated in Table~\ref{tab4}, ensemble is still inferior to D-PCN. More importantly, D-PCN is orthogonal to model ensemble just like HD-CNN~\cite{yan2015hd} and DaulNet~\cite{Hou2017DualNet}, ensemble of D-PCNs can further improve the performance.

\begin{table}[!tb]
	\caption{\textbf{Comparison between model ensemble and D-PCN on CIFAR-100. }Model ensemble deploys two identical CNN models with different initialization and takes averaged prediction as final result.
	}
	\begin{center}
		\footnotesize
		\setlength{\tabcolsep}{9pt}
		\begin{tabular}{|l|c|c|c|} 
			\hline
			\textbf{} & \textbf{NIN} & \textbf{ResNet-20} & \textbf{ResNeXt-29,8x64d} \\
			\hline
			base network  & 66.63\%  & 67.89\%  & 81.77\%  \\
			\hline
			\textbf{model ensemble}  & 68.94\%  & 70.01\%  & 83.03\% \\
			\hline
			\textbf{D-PCN}& \textbf{71.10\%} & \textbf{72.39\%}  & \textbf{84.59\%} \\
			\hline
		\end{tabular}
	\end{center}
	\label{tab4}
\end{table}

\textbf{Compared with doubling width } We take NIN for this experiment. After doubling number of channels directly, accuracy is improved to 68.89\%, still lower than 71.10\% of D-PCN. 

\textbf{Experiments on ensemble and doubling width demonstrate that the improvement of D-PCN is not merely from more parameters.}

\textbf{Where to feed the discriminator }We take ResNet-20 for this experiment. Original D-PCN sends features from block3 to discriminator as shown in Figure~\ref{fig3}, here we choose block2 and block1 instead. And it just gets 71.50\% and 68.78\% accuracy respectively, lower than 72.39\% of original one. Furthermore, we try to bring features from both block2 and block3 to discriminator through a convolutional layer, and we achieve 72.89\% accuracy, a little higher than 72.39\% of original D-PCN.

\textbf{How to aggregate features } Here we experiment on NIN and WRN-16-4. We replace concatenating with sum, and it achieves 70.27\%, 78.84\% for NIN and WRN respective, lower than 71.10\% and 80.19\% using concatenating.

\textbf{More subnetworks } We take NIN for this experiment. By adjusting the loss function in Section~\ref{sec3.3}, which is clarified in supplementary material, we can deploy three parallel networks in D-PCN. We get 71.97\% accuracy, a little higher than 71.10\% of original one.

\subsection{Classification Results on ImageNet}

In this experiment, we investigate D-PCN with ResNet-18~\cite{he2016deep} on ImageNet32x32~\cite{chrabaszcz2017downsampled}, NIN-ImageNet on ImageNet~\cite{deng2009imagenet}, to prove that D-PCN can generalize to more complex dataset. ImageNet32x32 is a downsampled version of ImageNet with 32x32 pixel per image. The applied ResNet-18 has same structure as ResNet for CIFAR, but numbers of channels keep pace with ResNet for ImageNet. The structure of NIN-ImageNet stays the same as in supplementary material of DualNet~\cite{Hou2017DualNet}. For ImageNet32x32, we only shift dataset to range from 0 to 1 and then zero-center the datasets. The results are shown in Table~\ref{tab5}. D-PCN attains 4.394\% and 9.45\% promotion in Top1 and Top5 accuracy on ImageNet32x32 respectively. 

\begin{table}[!tb]
	\caption{\textbf{Accuracy of D-PCN on ImageNet without data augmentation. }No data augmentation method is adopted except zero-centering preprocessing.
	}
	\begin{center}
		\footnotesize
		\setlength{\tabcolsep}{9pt}
		\begin{tabular}{|l|c|c|} 
			\hline
			\textbf{D-PCN} & \textbf{Top1 accuracy} &\textbf{Top5 accuracy} \\
			\hline
			\textbf{base ResNet-18}  & 45.738\%  & 59.78\%   \\
			\hline
			\emph{classifier of subnet1} & 45.732\% &     \\
			\hline
			\emph{classifier of subnet2} & 45.884\% &     \\
			\hline
			\emph{extra classifier} & \textbf{50.132\%} & \textbf{69.23\%}   \\
			\hline
		\end{tabular}
	\end{center}
	\label{tab5}
\end{table}

\begin{table}[tb]
	\caption{\textbf{Top 1 accuracy on ILSVRC-2012 ImageNet. }Both DualNet and D-PCN are based on NIN-ImageNet. The overall accuracy means accuracy of final prediction in DualNet and D-PCN on validation set.
	}
	\begin{center}
		\footnotesize
		\setlength{\tabcolsep}{10pt}
		\begin{tabular}{|l|c|c|} 
			\hline
			\textbf{}  &\textbf{DualNet}~\cite{Hou2017DualNet}  &\textbf{D-PCN} \\
			\hline
			\emph{base NIN-ImageNet}  & 59.15\%  & 58.94\%   \\
			\hline
			\textbf{Overall Accuracy} & 60.44\% &  \textbf{61.27\%}  \\
			\hline
		\end{tabular}
	\end{center}
	\label{tab5a}
\end{table}

For original ImageNet, the structure of NIN-ImageNet stays the same as in supplementary material of DualNet~\cite{Hou2017DualNet}. As presented in Table~\ref{tab5a}, D-PCN surpasses DualNet, and gains 2.33\% improvement versus base NIN-ImageNet.

For all above experiments, since philosophy of D-PCN is to coordinate two-stream networks to learn different representations, it's natural to use single CNN as baseline. And we can draw a conclusion that our work makes sense by introducing the novel framework and training strategy to compel two CNNs to learn distinctive features.

\subsection{Visualization}
\label{vi}
In experiments, loss from discriminator is getting quite small. We conjecture that in adversarial learning discriminator always wins, \emph{i.e.}, loss of the discriminator goes to very low fast. And the training rule in section~\ref{sec3.3} will make optimization of discriminator even easy.

\begin{figure}[tb]\centering
	\includegraphics[width=1\textwidth]{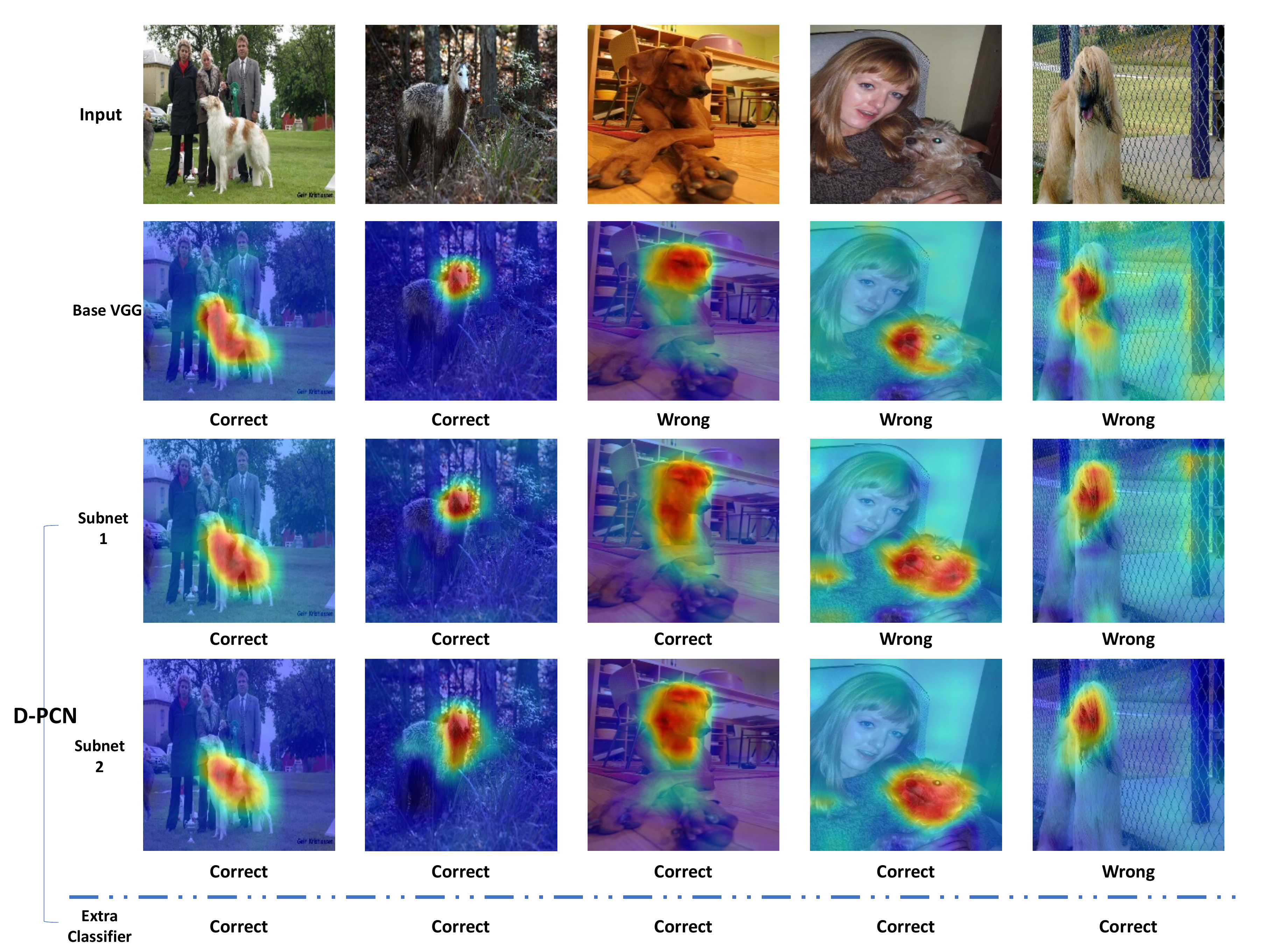}
	\caption{Grad-CAM visualization of VGG16 on Stanford Dogs. The correct or wrong means whether input is classified correctly by its own classifier. Last row represents results of extra classifier in D-PCN. There is no visualization for extra classifier since it takes in fused features.}
	\label{fig5}
\end{figure}

To confirm our point of view, we apply Grad-CAM~\cite{selvaraju2016grad} to give visual explanations for base VGG16 and two networks in D-PCN on Stanford Dogs~\cite{khosla2011novel} dataset, which possesses larger resolution better for visualization. Grad-CAM is extended from CAM~\cite{zhou2016learning} and is applicable to a wide variety of CNN models. Stanford Dogs~\cite{khosla2011novel} is a fine grained classification dataset, which has 120 categories of dogs and total 20580 images, where 12000 are for training. It's quite suitable for proving effectiveness of D-PCN since some different categories of dogs have very similar traits and are hard to be distinguished. D-PCN based on VGG16 is trained like other CNN models above. Here we select some representative pictures for visualization, as shown in Figure~\ref{fig5}, where red zone represents class-discriminative regions for network while blue zone is on the contrary. Like picture 4 in Figure~\ref{fig5}, original VGG only focuses on mouth, which maybe the reason of misclassification. Subnetworks localize more related areas and final result of D-PCN is correct.

\begin{figure}[tb]\centering
	\includegraphics[width=1\textwidth]{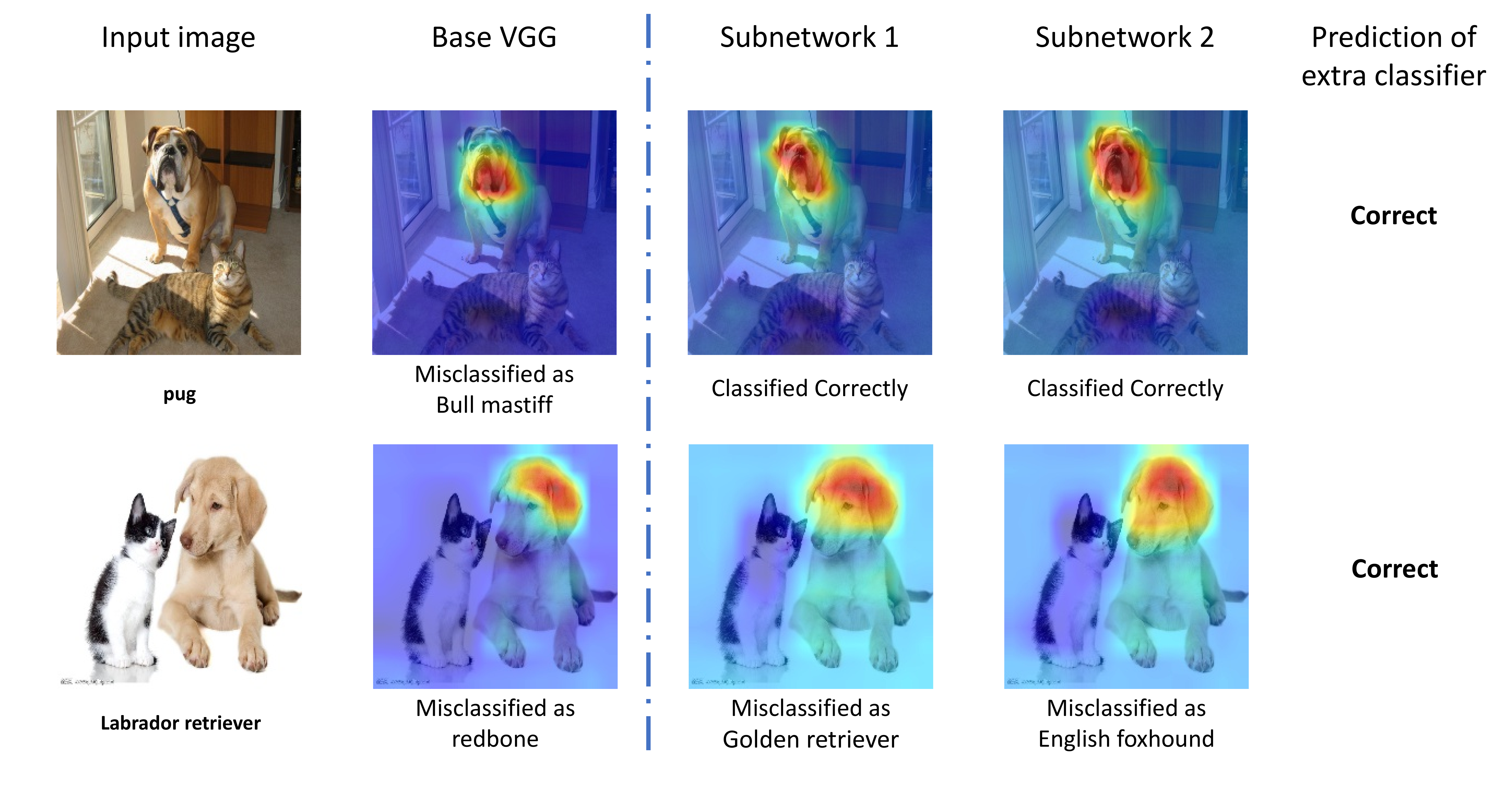}
	\caption{Grad-CAM visualization of VGG16 trained on Stanford Dogs. On the right of dotted line is visualization of D-PCN. Below pictures are predictions, correctness of final prediction from D-PCN is in the rightmost.}
	\label{fig6}
\end{figure}

And we select two images consisting of a cat, and a dog belonging to one category in Stanford Dogs, to test the models. Images are from internet. Visualization is shown in Figure~\ref{fig6}. We list category predictions of networks. That manifests the reason why CNNs misclassify some categories of dogs maybe some features vital to distinguish similar object get omitted, since network fails to observe some aspects. An extreme arresting discovery is that two networks even localize cat with no supervision information (Please pay attention to blue zone).

From these experiments, we can see that two subnetworks not only focus on different regions, but also localize more accurately than base network, which means discriminator does play a part in D-PCN even though loss from discriminator will be very small in training. We conjecture that there are always minor perturbations between subnetworks which make them diverse. Sometimes one of parallel networks predicts wrong or both misclassify objects, but extra classifier in D-PCN gives right answer eventually. Pleased noted like mentioned in Section~\ref{sec3.3} there are duplications among representations of subnets, but difference exists. This certifies the diversity (with redundance) of features from two-stream networks in D-PCN. By the way, accuracy of base VGG16 is 72.88\%, and accuracy of subnetworks and extra classifier are 73.36\%,73.53\% and 75.86\% respectively, which also signifies generalization of D-PCN on large dataset. Visualizations of different types can be found in supplementary material which further verifies our motivation.

\subsection{Segmentation on PASCAL VOC 2012}
Since D-PCN can coordinate parallel networks to learn different features, we think it can improve performance of other vision tasks. Here we put D-PCN into FCN-8s~\cite{long2015fully} on PASCAL VOC 2012 semantic segmentation task. ResNet-18 and ResNet-34 are chosen as base model. In training we set all $\lambda$ in Section~\ref{sec3.3} to 0.2, and extra classifier turns into convolutional layers corresponding to FCN. Two networks are initialized with pre-trained model on ImageNet. Experiment results are shown in Table~\ref{tab6}. D-PCN achieves 1.458\% and 1.304\% mIoU improvement respectively. Although improvements of testing mIoU are quite small, we found that training mIoU increases greatly. The results imply that convergence of networks rises significantly\footnote{We think it may explain why parallel networks can localize cat in Section~\ref{vi}, because subnetworks catch enough information to know what's dog.}.

\begin{table}[!tb]
	\caption{\textbf{Segmentation results on PASCAL VOC 2012. } We take FCN as base segmentation model. Training part of dataset is for training while validation part is for testing.
	}
	\begin{center}
		\footnotesize
		\setlength{\tabcolsep}{9pt}
		\begin{tabular}{|l|c|c|} 
			\hline
			\textbf{} & \textbf{ResNet-18} &\textbf{ResNet-34} \\
			\hline
			\hline
			\textbf{base model}  && \\
			\hline
			\emph{training mIoU of base model} & \textbf{69.139\%} &   \textbf{74.259\%}\\
			\hline
			\emph{testing mIoU of base model} & 50.352\% &  55.335\%\\
			\hline
			\hline
			\textbf{D-PCN}  & &  \\
			\hline
			\emph{training mIoU of subnetwork1} & \textbf{85.436\%} &   \textbf{87.557\%} \\
			\hline
			\emph{training mIoU of subnetwork2} & \textbf{85.341\%} & \textbf{87.647\%} \\
			\hline
			\emph{testing mIoU of subnetwork1} & 50.536\% & 56.026\%\\
			\hline
			\emph{testing mIoU of subnetwork2} & 50.601\% & 56.101\% \\
			\hline
			\emph{testing mIoU of D-PCN} & 51.810\% &  56.639\%\\
			\hline
		\end{tabular}
	\end{center}
	\label{tab6}
\end{table}

\section{Conclusion}
In this paper, we propose a novel framework named D-PCN to boost the performance of CNN. The parallel networks in D-PCN can learn discriminative and distinctive features via a discriminator. The fused features are more discriminative. An effective training method inspired by adversarial learning is introduced. D-PCNs based on various CNN models are investigated on CIFAR-100 and ImageNet datasets, and achieve promotion. In particular, it gets state-of-the-art performance on CIFAR-100 compared with related works. Additional experiments are conducted for visualization and segmentation. In the future, we will deploy D-PCN in other tasks efficiently, such as detection and segmentation.

\bibliographystyle{splncs}
\bibliography{D-PCN}

\newpage

{
	\large
	\textbf{Supplementary Material} 
}
\appendix
\section{More subnetworks } By adjusting the loss function in Section~\ref{sec3.3}, we can deploy three parallel networks in D-PCN.

Loss functions in training step 2 are defined as following.
\begin{eqnarray}
L_1=L_{cls_1}+\lambda L_{D_1}\\
L_{D_1}=\frac{1}{n}\sum^n{{[1-D(E_1(input))]^2}}
\end{eqnarray}
\begin{eqnarray}
L_2=L_{cls_2}+\lambda L_{D_2}\\
L_{D_2}=\frac{1}{n}\sum^n{{[D(E_2(input))]^2}}
\end{eqnarray}
\begin{eqnarray}
L_3=L_{cls_3}+\lambda L_{D_3}\\
L_{D_3}=\frac{1}{n}\sum^n{{[0.5-D(E_3(input))]^2}}
\end{eqnarray}
\begin{equation}
L_D=L_{D_1}+L_{D_2}+L_{D_3}
\end{equation}

Loss functions for subnetwork 2\&3 in training step 1 stay the same as in step 2. And $\lambda$ is set to 1.

\section{Visualizations for all images through Grad-CAM}
\begin{figure}[tb]\centering
	\includegraphics[width=0.85\textwidth]{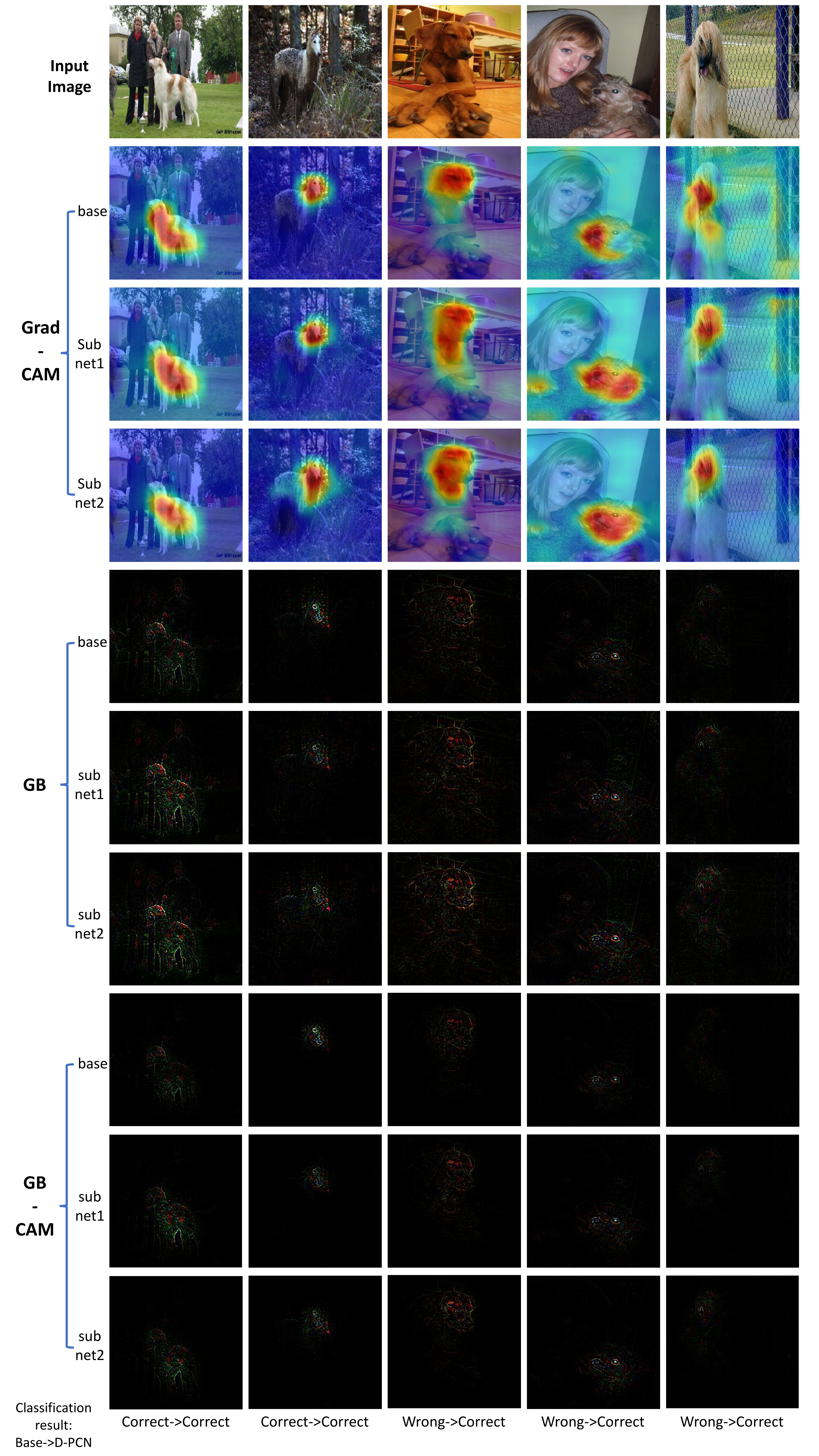}
	\caption{Grad-CAM visualization of VGG16 trained on Stanford Dogs. Best viewed in color/screen.}
	\label{fig7}
\end{figure}

\begin{figure}[tb]\centering
	\includegraphics[width=0.85\textwidth]{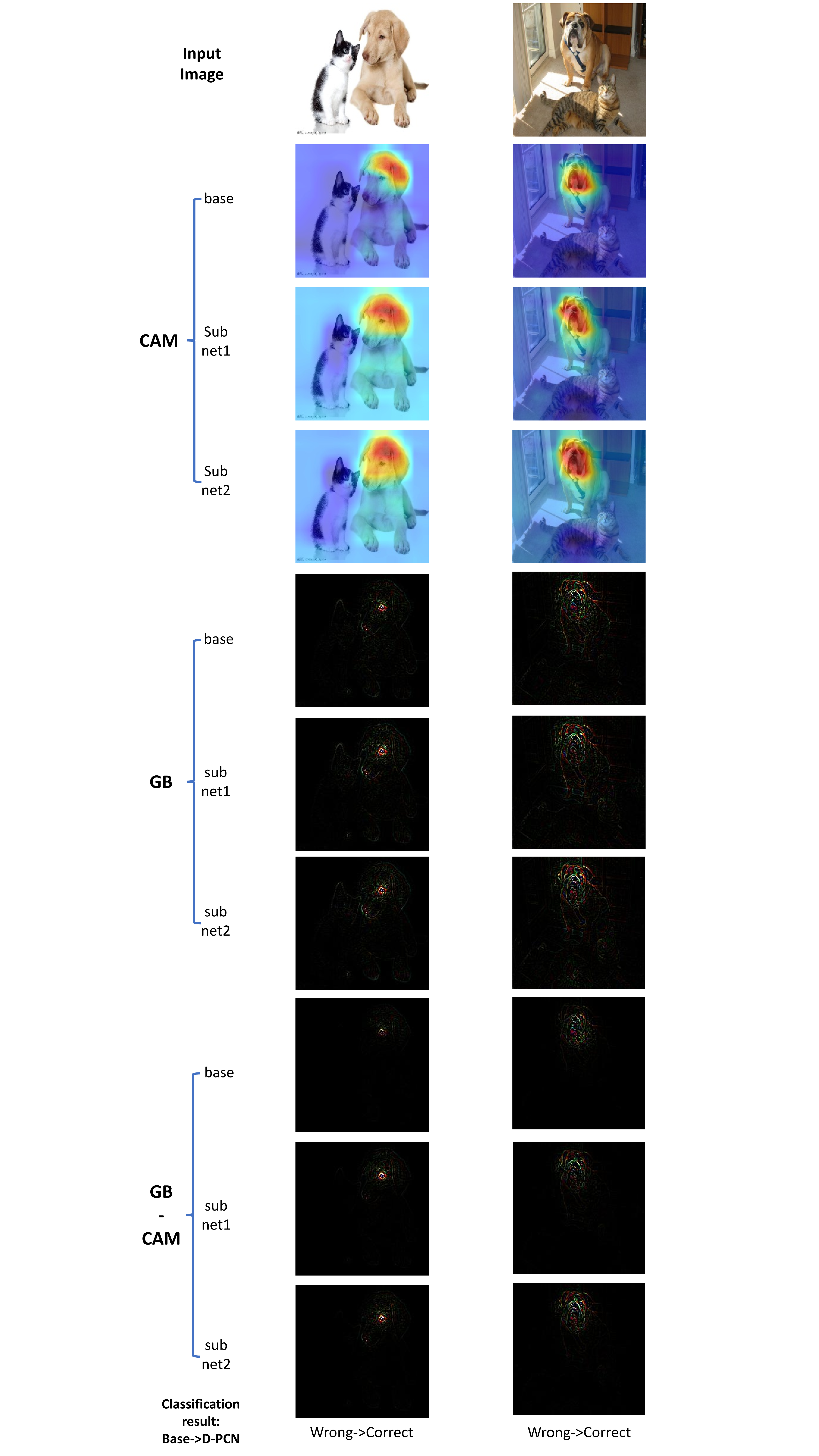}
	\caption{Grad-CAM visualization of VGG16 trained on Stanford Dogs. Best viewed in color/screen.}
	\label{fig8}
\end{figure}

The visualizations through Grad-CAM for all images are shown as following. GB means Guided Backpropagation, and GB-CAM means GB + Grad-CAM which is achieved by fusing GB and Grad-CAM. According to~\cite{selvaraju2016grad}, GB highlights all contributing features and GB-CAM can identify important features like stripes, pointy ears and eyes. Difference can be found between base VGG network and D-PCN in Figure~\ref{fig7} and Figure~\ref{fig8}, which demonstrates that features from two subnetworks in D-PCN are indeed diverse.

\end{document}